\documentclass{article}

\usepackage[utf8]{inputenc} % allow utf-8 input
\usepackage[T1]{fontenc}    % use 8-bit T1 fonts
\usepackage[colorlinks=true,linkcolor=black,anchorcolor=black,citecolor=black,filecolor=black,menucolor=black,runcolor=black,urlcolor=black]{hyperref}       % hyperlinks
\usepackage{url}            % simple URL typesetting
\usepackage{booktabs}       % professional-quality tables
\usepackage{amsfonts}       % blackboard math symbols
\usepackage{nicefrac}       % compact symbols for 1/2, etc.
\usepackage{microtype}     % microtypography
\usepackage{xcolor}         % colors
\usepackage{bm}         % italic-bold
\usepackage{graphicx}  		% figures
\usepackage{amsmath} % eqref
\usepackage{cite} % cite
\usepackage{diagbox}
\usepackage{siunitx}
\usepackage{multirow}
\usepackage{tabularx}
\usepackage{hhline}
\usepackage{arydshln}		% dotted lines in a table
\usepackage{xcolor}
\usepackage{mathtools} % coloneqq
\usepackage{amsthm}			% theorems
\usepackage{amssymb}			% varnothing
\usepackage{algorithm}
\usepackage{pifont}
\usepackage{threeparttable}
\usepackage{fullpage}
\usepackage[letterpaper,
            left=1.67cm,
            right=1.67cm,
            top=1.8cm,
            bottom=1.69in]{geometry}

\usepackage{algpseudocode}	
\usepackage{arydshln}
\usepackage[export]{adjustbox}
\usepackage{booktabs}

\usepackage{mathtools} 
\usepackage{microtype} 
\usepackage{multirow}
\usepackage{nicefrac}       
\usepackage{pifont}   

\usepackage{url}
\usepackage{siunitx}
\usepackage{tabularx}
\usepackage{threeparttable}
\usepackage{xcolor}  

\definecolor{darkblue}{rgb}{0,0 ,0.542}
\definecolor{lightgreen}{rgb}{.9,1,.9}
\definecolor{lightred}{rgb}{1,.415,.415}
\definecolor{lightblue}{rgb}{.415,.415,1}

\newcolumntype{L}[1]{>{\raggedright\arraybackslash}p{#1}}
\newcolumntype{C}[1]{>{\centering\arraybackslash}p{#1}}
\newcolumntype{R}[1]{>{\raggedleft\arraybackslash}p{#1}}

\theoremstyle{plain} % plain = italic, definition = roman

\newtheorem{theorem}{Theorem}

\newtheorem{assumption}{Assumption}

\def\defn{\,\coloneqq\,}

\def\Im{{\mathsf{Im}}}
\def\prox{{\mathsf{prox}}}

\def\R{\mathbb{R}}

\def\ebm{{\bm{e}}}

\def\xbm{{\bm{x}}}
\def\zbm{{\bm{z}}}
\def\ybm{{\bm{y}}}
\def\zbm{{\bm{z}}}
\def\sbm{{\bm{s}}}

\def\nbm{{\bm{n}}}

\def\Abm{{\bm{A}}}

\def\Ibm{{\bm{I}}}

\def\Abm{{\bm{A}}}

\def\Ibm{{\bm{I}}}

\def\Ncal{{\mathcal{N}}}

\def\Xcal{{\mathcal{X}}}

\def\Dsf{{\mathsf{D}}}

\def\Tsf{{\mathsf{T}}}

\def\Tsf{{\mathsf{T}}}

\def\Dsf{{\mathsf{D}}}

\def\xbmhat{{\widehat{\bm{x}}}}

\def\argmin{\mathop{\mathsf{arg\,min}}} % Argument of a minimization

\title{Convergence of Nonconvex PnP-ADMM with MMSE Denoisers}
\date{}

\author{
Chicago Park\textsuperscript{*}, Shirin Shoushtari\textsuperscript{*}, Weijie Gan, and Ulugbek S.\ Kamilov\\
\small Washington University in St. Louis, MO 63130, USA\\
\small \texttt{\{chicago,  s.shirin, weijie.gan, kamilov\}@wustl.edu}
}

\begin{document}

\maketitle
\let\thefootnote\relax\footnote{\textsuperscript{*}These authors contributed equally.}
\let\thefootnote\relax\footnote{This material is based upon work supported by the NSF CAREER award under grant CCF-2043134.}
%\thanks{This material is based upon work supported by the NSF CAREER award under grant CCF-2043134.}

\vspace{-3em} % Adjust as needed
\begin{abstract}
\medskip\noindent
Plug-and-Play Alternating Direction Method of Multipliers (PnP-ADMM) is a widely-used algorithm for solving inverse problems by integrating physical measurement models and convolutional neural network (CNN) priors. PnP-ADMM has been theoretically proven to converge for convex data-fidelity terms and nonexpansive CNNs. It has however been observed that PnP-ADMM often empirically converges even for expansive CNNs. This paper presents a theoretical explanation for the observed stability of PnP-ADMM based on the interpretation of the CNN prior as a minimum mean-squared error (MMSE) denoiser. Our explanation parallels a similar argument recently made for the iterative shrinkage/thresholding algorithm variant of PnP (PnP-ISTA) and relies on the connection between MMSE denoisers and proximal operators. We also numerically evaluate the performance gap between PnP-ADMM using a nonexpansive DnCNN denoiser and expansive DRUNet denoiser, thus motivating the use of expansive CNNs.

\end{abstract}

\section{Introduction}
\label{sec:intro}

A fundamental problem in computational imaging is the recovery of an unknown image $\xbm\in \R^n$ from noisy measurements 
$$\ybm = \Abm \xbm + \ebm,$$
where $\Abm \in \R^{m\times n}$ is the measurement operator and $\ebm \in \R^n$ is the measurements noise. It is common to formulate the recovery as a composite optimization problem
\begin{equation}
\label{Eq:optimization}
\xbmhat = \argmin_{\xbm\in\R^n} f(\xbm) \quad\text{with}\quad f(\xbm) = g(\xbm) + h(\xbm)\ ,
\end{equation}
where the function $g$  represents the data-fidelity term and $h$ denotes the regularizer or the prior term. 

\medskip\noindent
Proximal algorithms are commonly used to solve the optimization problems in eq.~\eqref{Eq:optimization} when the functions $g$ or $h$ are nonsmooth. For example, iterative shrinkage/thresholding algorithm (ISTA)~\cite{Figueiredo.Nowak2003, Beck.Teboulle2009} and alternating direction method of multipliers (ADMM)~\cite{Afonso.etal2010, Boyd.etal2011} have been widely-used in the context of imaging inverse problems. The iterations of ADMM can be expressed as
\begin{subequations}\label{Eq:InPnP-ADMM1}
\begin{align}
\begin{split} \label{Eq:xVarup1}
 \xbm^k \leftarrow \prox_{\gamma g} (\zbm^{k-1} - \sbm^{k-1})
\end{split}\\
\begin{split} \label{Eq:PnPsub1}
 \zbm^k \leftarrow \prox_{\gamma h}(\xbm^{k} + \sbm^{k-1})
\end{split}\\
\begin{split}\label{Eq:sUpdate1}
\sbm^k \leftarrow \sbm^{k-1} + \xbm^{k} - \zbm^{k}
\end{split}
\end{align}
\label{straincomponent1}
\end{subequations}
The key operation within ADMM is the \emph{proximal operator}
\begin{equation}
    \label{Eq:proximal}
    \prox_{\gamma h} (\zbm) := \argmin_{\xbm\in\R^n} \left\{\frac{1}{2}\|\xbm - \zbm\|_2^2 + \gamma h(\xbm)\right\},
\end{equation}
where the parameter $\gamma >0$ controls the influence of $h$. When $h(\xbm) = -\log\left ( p_\xbm(\xbm)\right )$, the proximal operator can be interpreted as a maximum a posteriori (MAP) estimator for the additive white Gaussian noise (AWGN) denoising problem
\begin{equation}
    \label{Eq:NoiseProb}
    \zbm = \xbm + \nbm \quad \text{with} \quad \xbm\sim p_\xbm, \quad \nbm \sim \Ncal(\bm{0}, \sigma^2 \Ibm).
\end{equation}
The statistical interpretation of the proximal operator as a denoiser has led to the development of the plug-and-play ADMM (PnP-ADMM)~\cite{Venkatakrishnan.etal2013}, where $\prox_{\gamma h}$ is replaced with a more general image denoiser $\Dsf_\sigma$
\begin{subequations}\label{Eq:InPnP-ADMM2}
\label{straincomponent2}
\begin{align}
\label{Eq:xVarup2}
&\xbm^k \leftarrow \prox_{\gamma g} (\zbm^{k-1} - \sbm^{k-1})\\
\label{Eq:PnPsub2}
&\zbm^k \leftarrow \Dsf_\sigma(\xbm^{k} + \sbm^{k-1})\\
\label{Eq:sUpdate2}
&\sbm^k \leftarrow \sbm^{k-1} + \xbm^{k} - \zbm^{k}, 
\end{align}
\end{subequations}
where $\sigma > 0$ controls the relative strength of the denoiser. The popularity of deep learning has led to a wide adoption of PnP for exploiting learned priors specified through convolutional neural networks (CNNs), leading to its state-of-the-art performance in a variety of applications~\cite{Zhang.etal2017a, Ahmad.etal2020, Zhang.etal2022}.

\begin{figure*}[t]
\centering
\includegraphics[width=18.3cm]{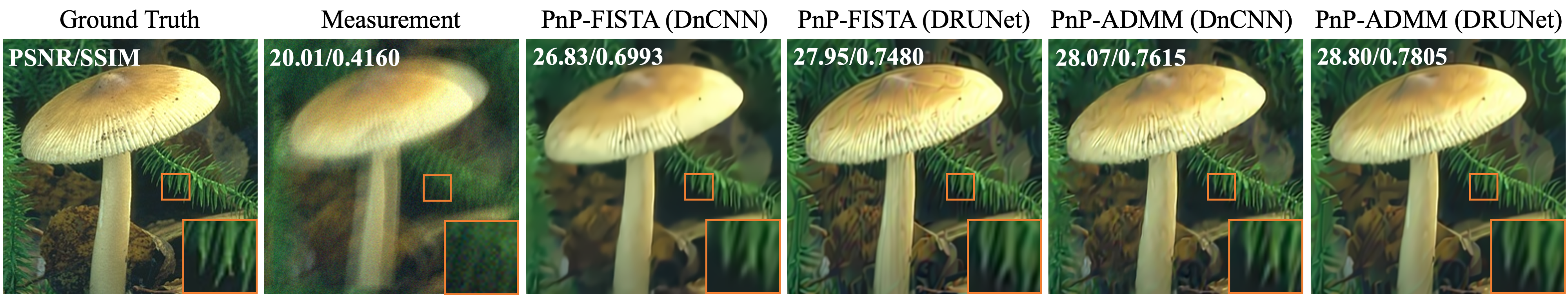}
   \caption{\small Comparison of four different methods for deblurring a color image with a noise level of 0.03. The reconstruction performance is quantified using PSNR and SSIM in the top-left corner of each image. Note the improved performance of PnP-ADMM using an expansive DRUNet denoiser compared to nonexpansive DnCNN denoiser. }
%\label{fig:short}
\label{architecture1}
\end{figure*}

\medskip\noindent
There has been significant interest in theoretically understanding the convergence behavior of PnP-ADMM~\cite{Sreehari.etal2016, Chan.etal2016, Ryu.etal2019, Reehorst.Schniter2019, Sun.etal2021, Hurault.etal2022a} (see also the recent review~\cite{Kamilov.etal2023}). The most well-known theoretical convergence results for PnP-ADMM are based on monotone operator theory, and require for $g$ to be convex and $\Dsf_\sigma$ or its residual to be nonexpansive~\cite{Sreehari.etal2016, Ryu.etal2019, Sun.etal2021}. 
Another well-known result does not assume nonexpansiveness by instead requiring bounded $\nabla g$ and $\Dsf_\sigma$~\cite{Chan.etal2016}. Recently PnP-ADMM was also analyzed for proximal operators associated with denoisers trained as gradients of explicitly specified regularizers~\cite{Hurault.etal2022a}. 

\medskip\noindent
While the current theory provides useful insights into the stability of PnP-ADMM iterations, it has been observed that PnP-ADMM often converges for denoisers that are expansive, unbounded, and/or are not trained using any explicitly specified function $h$. The goal of this paper is to offer an explanation for this stability by building on the recent analysis of PnP-ISTA for denoisers that perform minimum mean squared error (MMSE) estimation~\cite{Xu.etal2020}. The analysis of PnP-ISTA relies on an elegant connection between MMSE denoisers and proximal operators established by Gribonval~\cite{Gribonval2011}. What makes this connection pertinent is that the MSE is frequently used as a loss function for training state-of-the-art image denoisers. We explicitly relate the statistical interpretation of CNNs trained as MMSE denoisers with the nonconvex analysis of the traditional ADMM algorithm~\cite{li2018simple, wang2018convergence, guo2017convergence, jiang2019structured,yashtini2021multi}. We numerically motivate our theoretical exposition by comparing PnP-ADMM using two different pre-trained CNN denoisers, a nonexpansive DnCNN and an expansive DRUNet. The numerical results show the convergence of PnP-ADMM with an expansive denoiser and highlight the limitations of nonexpansive denoisers as priors within the PnP framework.

\begin{figure*}[t]
\begin{center}
\includegraphics[width=0.75\linewidth]{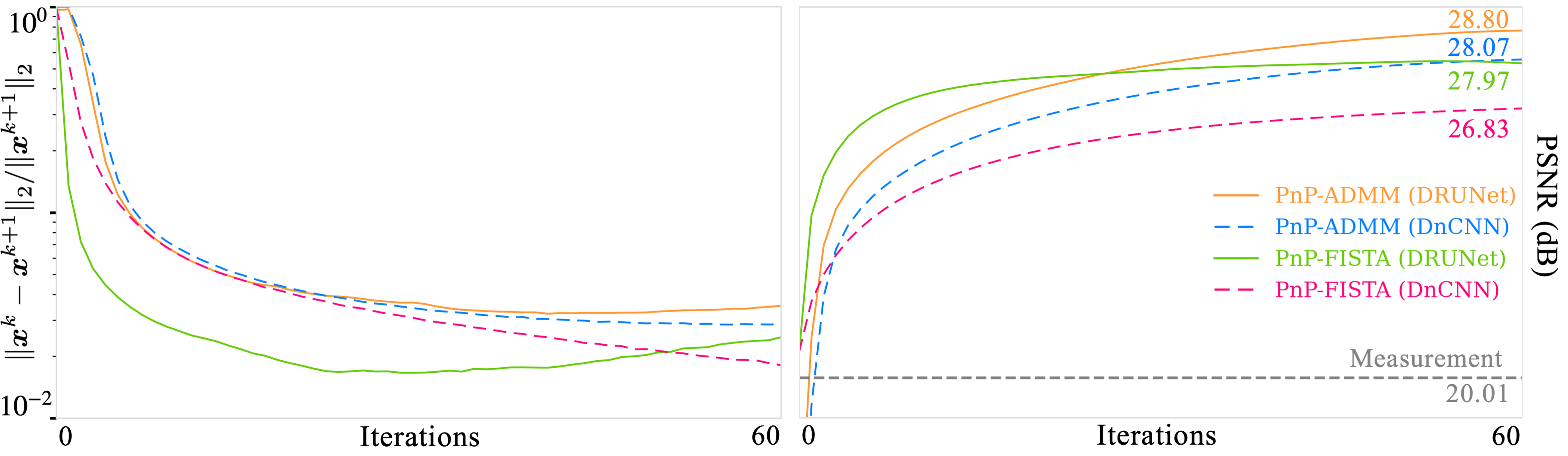}
\end{center}
    \caption{\small Comparison of PnP-ADMM and PnP-FISTA, each using a non-expansive DnCNN denoiser and an expansive DRUNet denoiser. The figure plots the evolution of $\|\xbm^{k} - \xbm^{k+1}\|_2 / \|\xbm^{k+1}\|_2$, while the right one that of PSNR (dB).}
%\label{fig:short}
\label{architecture2}
\end{figure*}

\section{Convergence Theory}
\label{sec:Theo}

We state three assumptions required to establish the convergence of PnP-ADMM under a nonconvex data-fidelity term $g$ and a MMSE denoiser $\Dsf_\sigma$. 
\begin{assumption}
    \label{As:NonDegPrior}
    The prior $p_\xbm$ is non-degenerate over $\R^n$. 
\end{assumption}

\noindent
We call the probability distribution $p_\xbm$ a degenerate distribution over $\R^n$, if its support lies on a space with lower dimension than $n$. For the MMSE denoiser $\Dsf_\sigma$ with the image set of $\mathcal{X} \defn \Im (\Dsf_\sigma)$, Assumption~\ref{As:NonDegPrior} is required to establish an explicit link between $\Dsf_\sigma$ and the following regularizer~\cite{Gribonval2011}
\begin{equation}
\label{Eq:ExpReg}
h_{\textsf{\tiny mmse}}(\xbm) \defn \begin{cases}
-\frac{1}{2\gamma}\|\xbm - \Dsf_{\sigma}^{-1}(\xbm)\|_2^2 + \frac{\sigma^2}{\gamma} h_{\sigma}(\Dsf_{\sigma}^{-1}(\xbm)) & \text{for } \xbm \in \mathcal{X} \\
+\infty & \text{for } \xbm \notin \mathcal{X},
\end{cases}
\end{equation}
where $\gamma > 0 $ is the step-size, $\Dsf_\sigma^{-1}$ is a smooth inverse mapping over $\mathcal{X}$, and $h_\sigma (\cdot) = - \log \left ( p_{\zbm}(\cdot)\right )$, where $p_\zbm$ os the probability distribution in~\eqref{Eq:NoiseProb}. Due to smoothness of both $\Dsf_\sigma^{-1}$ and $h_\sigma$, the function $h_{\textsf{\tiny mmse}}$ is smooth for all $\xbm \in \mathcal{X}$. For additional discussion see~\cite{Xu.etal2020,Gribonval2011}. 
\begin{assumption}
    \label{As:LipG&H}
    Function $g$ is continuously differentiable. The function $h$ in~\eqref{Eq:ExpReg} associated with $\Dsf_\sigma$ has a Lipschitz continuous gradients with constants $M>0$ over the set $\mathcal{X}$. 
\end{assumption}
\noindent
The smoothness of $g$ is a mild and commonly used assumption~\cite{jiang2019structured}. Since it is known (see~\cite{Gribonval2011, Xu.etal2020}) that the implicitly defined function $h$ is infinitely differentiable over $\Xcal$, our assumption is a mild extension that additionally requires Lipschitz continuity of the gradient.
\begin{assumption}
    \label{As:BoundedF}
    The explicit data-fidelity term and the implicit regularizer are bounded from below
\begin{equation*}
\inf_{\xbm \in \R^n} g(\xbm) > -\infty, \quad \inf_{\xbm \in \R^n} h(\xbm) > -\infty.
\end{equation*}
\end{assumption}
\noindent
Assumption~\ref{As:BoundedF} implies that there exists $f^\ast > -\infty$ such that $f(\xbm) \geq f^\ast$ for all $\xbm \in \R^n$.
\begin{theorem}\label{theorem1}
Run PnP-ADMM with a MMSE denoiser under Assumptions~\ref{As:NonDegPrior}-\ref{As:BoundedF} using a fixed step-size $0<\gamma\leq 1/(2M)$.  Then, then the iterates satisfy $\|\nabla f(\xbm^k)\|_2 \to 0$ as $k \to \infty$, where $f = g+h$ with $h$ defined in~\eqref{Eq:ExpReg}.
\end{theorem}
\noindent
The proof is provided in the appendix. Theorem~\ref{theorem1} establishes the convergence of PnP-ADMM with MMSE denoisers to a critical point of the problem~\eqref{Eq:optimization} where $h$ is defined in~\eqref{Eq:ExpReg}. It is important to note that the proof does not require the convexity of $g$ or $h$, or nonexpansiveness of the denoiser. This result suggests the stability of PnP-ADMM under possibly expansive denoisers trained to minimize MSE. The convexity of $h_{\textsf{\tiny mmse}}$  requires for $\Dsf_\sigma$ to be monotone and nonexpansive, which does not hold for many CNNs. 

\section{Numerical Evaluation}
\label{sec:Num}

\medskip\noindent
We explore the convergence and performance of two PnP algorithms, PnP-ADMM and PnP-FISTA, on the problem of image deblurring using two popular CNN priors, DnCNN and DRUNet, trained to minimize mean squared error. The DnCNN is trained to be nonexpansive by removing its batch normalization layers and using spectral normalization to bind its Lipschitz constant. The evaluation of these methods is conducted using four different blur kernels and four CBSD68 test images at three distinct noise levels (0.01, 0.02, and 0.03). The performance is quantified using peak signal-to-noise ratio (PSNR) in dB and structural similarity index (SSIM) metrics.

\medskip\noindent
Figure 1 compares all four methods applied to a color image with a noise level of 0.03. It is observed that PnP-ADMM converges for the expansive denoiser and achieves significantly better results in terms of both PSNR and SSIM. PnP-FISTA with DRUNet outperforms the one using a nonexpansive DnCNN.
Figure 2 presents the convergence profiles of PnP-ADMM and PnP-FISTA with DnCNN and DRUNet as image priors. Theoretical support for the convergence of PnP-ADMM is established through Theorem~1, which does not impose any assumptions on the expansiveness of the image prior. The results in Figure 2 plot the evolution of $\|\xbm^k - \xbm^{k+1}\|_2 /\|\xbm^{k+1}\|_2 $ and PSNR for the image presented in Figure 1, supporting the theoretical observation that PnP-ADMM can converge for expansive CNNs.

\medskip\noindent
Table 1 presents the performance comparison of PnP-ADMM and PnP-FISTA for both image priors. The reported results are obtained by averaging PSNR values over all the test images and kernels, illustrated in Figure 3.
Note how PnP-ADMM outperforms PnP-FISTA in terms of PSNR for both priors. Additionally, the use of the expansive denoiser, DRUNet, results in better performance compared to the nonexpansive denoiser, DnCNN, regardless of the algorithm used. The presented findings emphasize the notable advantage that PnP methods can leverage advanced denoisers, resulting in enhanced performance.

\begin{table}[t]
\centering
\scalebox{0.9}{
\begin{tabular}{lcccccc}
\hline
\\[-2ex]
  &   \multicolumn{3}{c}{Noise Level}        \\ \cline{2-4} 
    \\[-2ex]
 PnP Methods   & \multicolumn{1}{c}{0.01} & \multicolumn{1}{c}{0.02} & \multicolumn{1}{c}{0.03}  && \multicolumn{1}{c}{avg}\\ \cline{3-7} 

\hline
\\[-2ex]
ADMM (DRUNet) &  \textbf{30.67} & \textbf{29.48} & \textbf{28.03}               && \textbf{29.39}  \\
\\[-2ex]
ADMM (DnCNN)  &  28.84 & 28.01 & 26.25               && 27.70
\\[-2ex]
\\ \cdashline{1-6}
\\[-2ex]

\\[-2ex]
FISTA (DRUNet) &  27.30 & 27.20 & 26.84               && 27.11
 \\
\\[-2ex]
FISTA (DnCNN)  &  26.57 & 26.40 & 25.90               && 26.29
 \\
\\[-2ex]

\hline
\end{tabular}}
\caption{\small Performance of PnP-ADMM and PnP-FISTA using two priors on image deblurring at different levels of noise.} 
\label{tab:static_error_table}
\end{table}

\begin{figure}[t]
\begin{center}
\includegraphics[width=.45\linewidth]{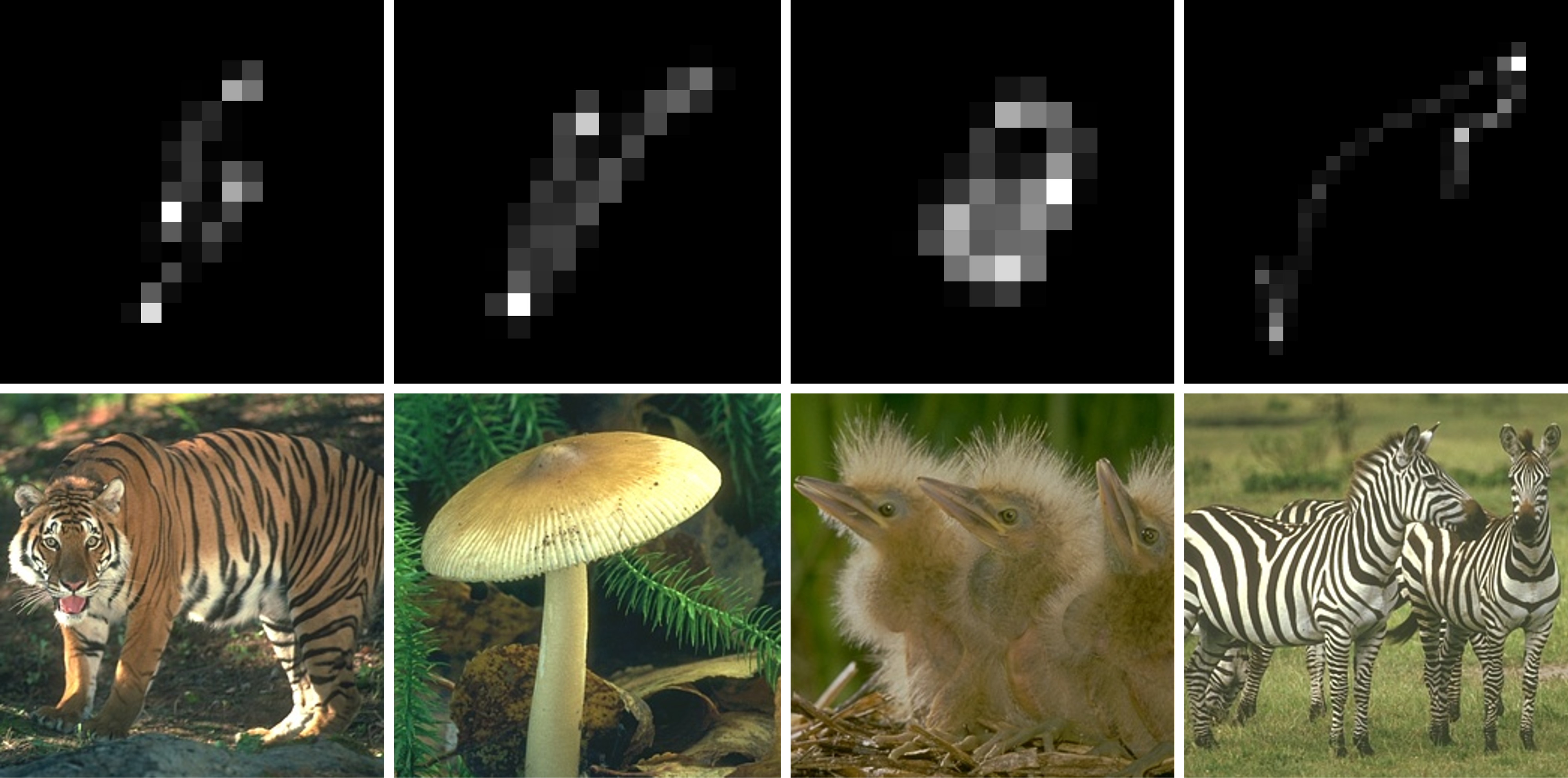}
\end{center}
   \caption{\small Four blur kernels \cite{Levin.etal2009} and test color images \cite{Martin.etal2001} used for the numerical evaluation.}
%\label{fig:short}
\label{architecture3}
\end{figure}

\section{Conclusion}
\label{sec:Con}
This paper presents new insights to the well-established PnP methodology by revisiting the convergence of PnP-ADMM under MMSE denoisers. We present convergence analysis of PnP-ADMM for expansive image denoisers and possibly nonconvex data-fidelity terms. We present numerical results highlighting the potential benefit of using expansive denoisers corresponding to CNNs trained to minimize MSE. Our findings emphasize that superior performance of PnP-ADMM under expansive denoisers can still come with stable convergence. This highlights the potential of combining state-of-the-art denoising techniques with the flexibility of PnP.

\appendix
\section{Appendix}

We now present the convergence analysis of PnP-ADMM without any convexity or nonexpansiveness assumptions.
\noindent
Consider the augmented Lagrangian of the objective function $f(\xbm) = g(\xbm) + h(\xbm)$
\begin{equation}
    \label{Eq:AugmentedLagrangian} \mu (\xbm, \zbm, \sbm )= g(\xbm) + h(\zbm) + \frac{1}{\gamma} \sbm^\Tsf (\xbm  -\zbm) +\frac{1}{2\gamma} \|\xbm - \zbm\|_2^2.
\end{equation}
For iteration $k\geq 1$ we have
\begin{equation}
    \label{Eq:LagrangInequality1}
    \mu (\xbm^k , \zbm^k , \sbm^k ) - \mu (\xbm^k , \zbm^k , \sbm^{k-1} )= \frac{1}{\gamma} \|\sbm^k - \sbm^{k-1}\|_2^2, 
\end{equation}
where we used the definition of the augmented Lagrangian in~\eqref{Eq:AugmentedLagrangian} and the ADMM update~\eqref{Eq:sUpdate2}. The results from~\cite{Gribonval2011} imply that for a MMSE denoiser we have $\Dsf_\sigma = \prox_{\gamma h}$. Therefore, from the optimality of proximal operator and the ADMM update~\eqref{Eq:sUpdate2}, we have $\sbm^k = \gamma \nabla h(\zbm^k)$. Using the Lipschitz continuity of $\nabla h$ in Assumption~\ref{As:LipG&H}, we have
\begin{equation}
    \label{Eq:LipofGradH}
    h(\zbm^k) - h(\zbm^{k-1}) 
    \leq \frac{1}{\gamma} (\sbm^k)^\Tsf (\zbm^k - \zbm^{k-1}) + \frac{M}{2}\|\zbm^k - \zbm^{k-1}\|_2^2. 
\end{equation}
From the fact that $\xbm^k = \prox_{\gamma g}\left(\zbm^{k-1} - \sbm^{k-1}\right )$, we have
\begin{equation}
    \label{Eq:LagrangInequality2}
    \mu (\xbm^{k} , \zbm^{k-1} , \sbm^{k-1} ) \leq  \mu (\xbm^{k-1} , \zbm^{k-1} , \sbm^{k-1} ).
\end{equation}
By combining equations~\eqref{Eq:LagrangInequality1},~\eqref{Eq:LipofGradH} , and~\eqref{Eq:LagrangInequality2}, we get  
\begin{align}
    \label{Eq:LagrangInequality3}
    \nonumber\mu (\xbm^k , \zbm^k , \sbm^k ) - \mu (\xbm^{k-1} , \zbm^{k-1} , \sbm^{k-1} )&\leq \frac{1}{\gamma} \|\sbm^k - \sbm^{k-1}\|_2^2 - \left(\frac{1-\gamma M}{2 \gamma}\right)\|\zbm^k - \zbm^{k-1}\|_2^2\\
    &\leq -\eta\|\zbm^k - \zbm^{k-1}\|_2^2, 
\end{align}
where $\eta \defn (1 - \gamma M - 2\gamma^2 M^2)/(2\gamma) $, and in the last inequality, we used 
\begin{equation} \label{Eq:BoundofS}
    \|\sbm^k - \sbm^{k-1}\|_2 = \gamma \|\nabla h(\zbm^k) - \nabla h(\zbm^{k-1})\|_2\leq \gamma M \|\zbm^k - \zbm^{k-1}\|_2.
\end{equation}
Since $0< \gamma \leq 1/(2M)$, we have $\eta > 0$, which implies that the augmented Lagrangian is monotonically decreasing. Using the fact that $\sbm^k = \gamma \nabla h(\zbm^k)$, we can write 
\begin{equation*}
    \mu (\xbm^k , \zbm^k , \sbm^k ) = g(\xbm^k) + h(\zbm^k) + \nabla h(\zbm^k)^\Tsf (\xbm^k - \zbm^k)+\frac{1}{2\gamma} \|\xbm^k - \zbm^k\|_2^2 \geq g(\xbm^k) + h(\xbm^k), 
\end{equation*}
where we used Lipschitz continuity of $\nabla h$ and the fact that $\gamma M \leq 1$. This inequality establishes that augmented Lagrangian is bounded from below due to the fact that both function $g$ and $h$ are bounded from below. This implies that there exists $\mu^* > -\infty$ such that almost surely $\mu^* \leq \mu(\xbm^k , \zbm^k , \sbm^k ), k\geq 1$. Thus, the augmented Lagrangian converges due to monotonicity. By summing both sides of eq.~\eqref{Eq:LagrangInequality3}  over $t\geq 1$ iteration, we have
\begin{equation}
    \label{Eq:convergenceIters}
   \sum_{k=1}^t \|\zbm^k - \zbm^{k-1}\|_2^2 \leq \frac{\mu(\xbm^0 , \zbm^0 , \sbm^0 ) -\mu(\xbm^t , \zbm^t, \sbm^t)}{\eta}
    \leq \frac{\mu(\xbm^0 , \zbm^0 , \sbm^0 ) -\mu^*}{\eta}, 
\end{equation}
which implies that $\|\zbm^k - \zbm^{k-1}\|_2^2 \rightarrow 0$ as $k \rightarrow \infty$. Eq.~\eqref{Eq:BoundofS} ensures convergence of  $\|\sbm^k - \sbm^{k-1}\|_2$ and  $\|\xbm^k - \zbm^{k}\|_2$ to $0$ as $k\rightarrow \infty$. For the objective function in~\eqref{Eq:optimization}, by adding and subtracting required terms, we have
\begin{align}
    \nonumber\|\nabla f(\xbm^k )\|_2 &=\|\nabla g(\xbm^k) + \nabla h(\xbm^k)\|_2 \\&= \|\nabla g(\xbm^k) + \frac{1}{\gamma} \left ( \xbm^k - \zbm^{k-1}+ \sbm^{k-1}\right )  +\nabla h(\xbm^k) +  \frac{1}{\gamma} \left ( \zbm^k  - \xbm^k - \sbm^{k-1}\right ) + \frac{1}{\gamma} (\zbm^{k-1}- \zbm^k)\|_2\\
    \nonumber& = \|\frac{1}{\gamma} (\zbm^{k-1}- \zbm^k) + \nabla h(\xbm^k) - \nabla h(\zbm^k)\|_2\\
    &\leq \frac{1}{\gamma}\|\zbm^{k}- \zbm^{k-1}\|_2 + M \|\xbm^k - \zbm^k\|_2,
\end{align}
where we used optimality of proximal operator from eq.~\eqref{Eq:xVarup2} and~\eqref{Eq:PnPsub2}, and in the last line, we used triangle inequality and Lipschitz continuity of $\nabla h$. From convergence of $\|\zbm^k - \zbm^{k-1}\|_2$ and $\|\xbm^k - \zbm^{k}\|_2$ to $0$ as $k \to \infty$, we get 
\begin{equation*}
    \|\nabla f(\xbm^k)\|_2 \to 0, 
\end{equation*}  
as $k \to \infty$. This shows that PnP-ADMM with MMSE denoiser converges to a critical point of the original objective function $f(\xbm)$. 


\bibliographystyle{IEEEbib}
\bibliography{refs}
\end{document}